\documentclass[10pt,twocolumn,letterpaper]{article}

\usepackage{iccv}
\usepackage{times}
\usepackage{epsfig}
\usepackage{graphicx}
\usepackage{amsmath}
\usepackage{amssymb}

\usepackage{tikzducks}
\usepackage{duckuments}

\usepackage{booktabs}
\usepackage{bm}
\usepackage{pifont}
\usepackage[inline]{enumitem}
\usepackage{soul}

\usepackage{dblfloatfix}
\usepackage{siunitx}
\usepackage{balance}

\usepackage{multirow}
\usepackage{makecell}

\usepackage[pagebackref=true,breaklinks=true,letterpaper=true,colorlinks,bookmarks=false]{hyperref}

\iccvfinalcopy % *** Uncomment this line for the final submission

\def\iccvPaperID{28} % *** Enter the ICCV Paper ID here

\ificcvfinal\fi

\def \ie {\emph{i.e.},}
\def \eg {\emph{e.g.},}
\def \etal {\emph{et al.}}

\newcommand{\tit}[1]{\smallbreak\noindent\textbf{#1.}}

\usepackage[capitalize]{cleveref}
\crefname{section}{Sec.}{Secs.}
\Crefname{section}{Section}{Sections}
\Crefname{table}{Table}{Tables}
\crefname{table}{Tab.}{Tabs.}

\begin{document}

%%%%%%%%% TITLE
\title{Volumetric Fast Fourier Convolution \\ for Detecting Ink on the Carbonized Herculaneum Papyri}

\author{Fabio Quattrini, Vittorio Pippi, Silvia Cascianelli, Rita Cucchiara\\
University of Modena and Reggio Emilia\\
Via Pietro Vivarelli, 10, Modena (Italy)\\
{\tt\small \{name.surname\}@unimore.it}
}
\maketitle
\ificcvfinal\thispagestyle{empty}\fi

%%%%%%%%% ABSTRACT
\begin{abstract}   
Recent advancements in Digital Document Restoration (DDR) have led to significant breakthroughs in analyzing highly damaged written artifacts. Among those, there has been an increasing interest in applying Artificial Intelligence techniques for virtually unwrapping and automatically detecting ink on the Herculaneum papyri collection. This collection consists of carbonized scrolls and fragments of documents, which have been digitized via X-ray tomography to allow the development of ad-hoc deep learning-based DDR solutions. In this work, we propose a modification of the Fast Fourier Convolution operator for volumetric data and apply it in a  segmentation architecture for ink detection on the challenging Herculaneum papyri, demonstrating its suitability via deep experimental analysis. To encourage the research on this task and the application of the proposed operator to other tasks involving volumetric data, we will release our implementation (\href{https://github.com/aimagelab/vffc}{https://github.com/aimagelab/vffc}).
\end{abstract}

%%%%%%%%% BODY TEXT
\section{Introduction}
\label{sec:introduction}
Some of the most valuable sources of information we have about ancient cultures and populations are the manuscripts and, in general, the artifacts with writings and pictures that survived history~\cite{qureshi2019hyperspectral,cascianelli2022lam,maarand2022comprehensive,pippi2023choose}. For this reason, even the smallest of such objects is precious for its potentially unique and impactful content.
Due to the fragility of the medium and their long history, most of the ancient manuscripts found by archaeologists can be extremely degraded and irreversibly damaged, and thus, challenging to handle and analyze. 
These challenges are even more critical for those ancient documents that were written on scrolls, which also necessitate being unrolled for reading.

Digital Document Restoration (DDR) aims to provide access to these kinds of documents and has therefore gained great interest from researchers and practitioners~\cite{doncescu1997former, brown2001document, pilu2001undoing, cao2003cylindrical, zhang2004restoration, gatos2007segmentation}. 
Among the DDR techniques, Virtual Unwrapping can be applied when the textual content is physically unreachable, as in the case of fragile scrolls.
Starting from a digital 3D volumetric representation obtained with X-ray micro-computed tomography~\cite{kennedy2006use}, this method entails reconstructing the 2D text image, allowing researchers, scholars, and the general public to visually inspect and study these historical artifacts. Early works on the application of Virtual Unwrapping were limited to specific use cases and entailed semi-manual pipelines~\cite{seales2004digital, lin2005opaque}. Later, fully automated digital unwrapping was applied to undamaged or partially damaged scrolls made of parchments~\cite{samko2014virtual, stromer2017browsing, stromer2018browsing, rosin2018virtual}, bamboo~\cite{stromer2018non, stromer2019virtual}, papyri \cite{allegra2016x, allegra2015virtual}, silver~\cite{hoffmann2015revealing, baum2021revisiting}, and the famous En-Gedi scroll~\cite{seales2016damage}, charred and charcoaled by a fire in early middle ages. 

\begin{figure}[t]
    \centering
    \includegraphics[width=0.47\textwidth]{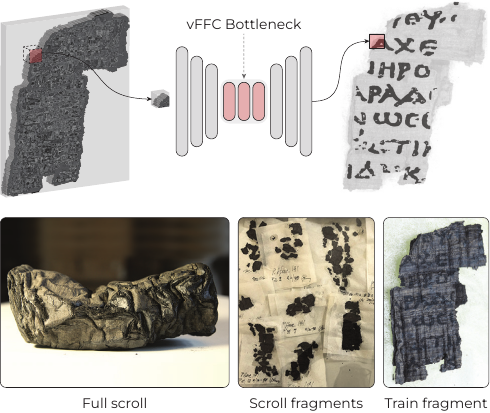}
    \caption{We propose an architecture featuring our devised volumetric Fast Fourier Convolution operator for the ink detection task on the fragments of carbonized Herculaneum papyri (parts of this image are from the official project website \href{https://scrollprize.org/}{https://scrollprize.org/}).}
    \label{fig:overview}
\end{figure}

When discussing ancient irreparably damaged scrolls, few examples carry as much significance and present as many challenges as the Herculaneum papyri. This collection of more than 1800 manuscripts, recovered from the Herculaneum Villa of the Papyri, presents a particularly important case due to its unique preservation state and the special interest of scholars~\cite{seabrook2015invisible}. 
Carbonized and buried in ashes during the Vesuvius eruption of 79 C.E., they represent the only intact surviving library from antiquity discovered in its original location~\cite{seabrook2015invisible}. Among the posed challenges, the ink is carbon-based, different from the metal-based ink in the En-Gedi scroll. Thus, the response of the ink to X-rays is not seemingly distinguishable from that of the carbonized papyrus support, making traditional Virtual Unwrapping ineffective. 

In the sight of this, in 2023, Parsons~\etal~\cite{parsons2023educelab} proposed to push forward DDR on this challenging collection by resorting to modern Computer Vision, Document Analysis, and Artificial intelligence techniques. To this end, they developed and released EduceLab-Scrolls, an open dataset containing volumetric scans of rolled scrolls and detached fragments from the Herculaneum papyri, which is the object of a dedicated ongoing competition\footnote{https://scrollprize.org/}. The goal is to develop advanced Virtual Unwrapping strategies to unroll these challenging scrolls and then perform ink detection on the unrolled sheets. Meanwhile, the fragments can be used to develop ink detection algorithms. In fact, for the fragments, it has been possible to obtain infrared images and, thus, ink maps that can be used for training supervised deep learning models. Thanks to their effort, this novel task on such challenging data is receiving increasing interest from the Artificial Intelligence community.

In this respect, we propose to tackle the task via an efficient, fully-Convolutional model featuring blocks inspired by Fast Fourier Convolutions (FFCs)~\cite{chi2020fast}, which we modify to handle volumetric data (Figure~\ref{fig:overview}). Note that the FFC operator exploits the spectral information of the input to expand its receptive field in the frequency domain and to handle pseudo-periodic patterns. Spatial FFC has been employed for tasks on 2D images but, to the best of our knowledge, it has never been applied to volumetric data. For this reason, we propose a modification to the original FFC operation that makes it able to handle volumetric data and thus be applied to the Herculaneum papyrus fragments scans for ink detection. Through a deep evaluation analysis, we provide useful insights on the challenging task of ink detection on Herculaneum papyri and on the proposed approach, which we demonstrate to be suitable for the task.
\section{Related Work}
\label{sec:related}
\tit{Digital Document Restoration} 
DDR encompasses Digital Imaging, Image Processing, and Computer Vision techniques for non-invasive content recovery from severely damaged ancient documents~\cite{doncescu1997former, brown2001document, pilu2001undoing, cao2003cylindrical, zhang2004restoration, gatos2007segmentation}. One of these techniques is Virtual unwrapping~\cite{seales2017reading}, which is mostly applied to documents that are too fragile to handle and analyze, as in the case of scrolls. 
First, the entire scroll is scanned, typically with X-ray micro-computed tomography (micro-CT)~\cite{kennedy2006use}. Then, each scroll sheet in the volumetric representation is segmented and projected into a 2D image. Depending on the specific use case, an additional texturing step (such as ink detection) can be performed. Early works operated on small contrived samples with semi-manual pipelines~\cite{seales2004digital, lin2005opaque}. The first fully-automated solution was proposed by Samko~\etal~\cite{samko2014virtual}, who developed a novel graph cut method for parchment scrolls sheets segmentation. In~\cite{hoffmann2015revealing, baum2021revisiting}, the authors read the contents of a damaged silver scroll by exploiting the specific characteristics of the material, such as the engraved and ruled text. Finally, the carbonized En-Gedi scroll was read in 2015 by Seales~\etal~\cite{seales2016damage}. 

\tit{Herculaneum papyri}
This work focuses on DDR of the Herculaneum papyri, a document collection that poses unprecedented challenges~\cite{seales2013virtual} to the application of a classical virtual unwrapping pipeline. 
Firstly, the papyrus layers have been compressed, crumpled, and deformed by the carbonization process, making sheet segmentation not straightforward. Secondly, the text is mostly written using carbon-based ink, which is almost invisible on the carbonized papyrus support in the X-ray scans. Thus, traditional virtual unwrapping, which heavily relies upon the visibility of ink or simple layers structure, is not applicable.  
In a recent work, Parker~\etal~\cite{parker2019invisibility} proved the detectability of carbon-based ink in micro-CT scans of the Herculaneum papyri using a 3D Convolutional Neural Network (Conv)~\cite{ji20123d} trained on subvolumes to detect ink in the central voxel. A further step towards DDR of Herculaneum papyri is due to Parsons~\etal~\cite{parsons2023educelab}, who proposed an open dataset containing volumetric scans of such collection. 
In this work, we tackle the problem of ink detection on the surface volumes of Herculaneum papyri fragments.

\begin{figure*}
    \centering
    \includegraphics[width=\textwidth]{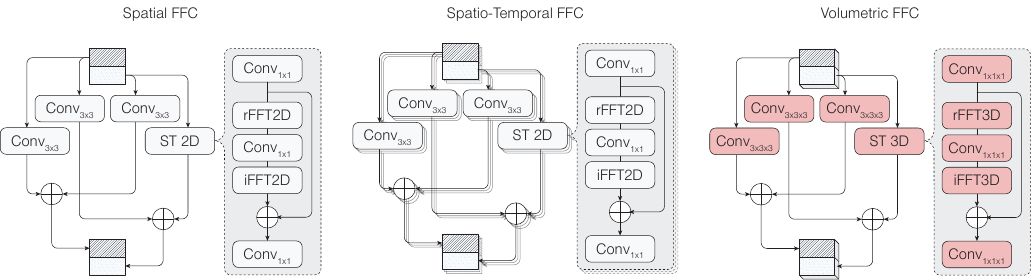}
    \caption{The standard spatial FFC (left) can be adapted to work on spatio-temporal data by replicating it along the channel dimension (center). Our proposed volumetric FFC (right) is designed to handle 3D volumetric data directly by combining 3D convolutions and 3D FFTs. For simplicity, we omit the batch normalization and ReLU operations in the schemes.}
    \label{fig:model}
\end{figure*}

\tit{Spectral Analysis}
Previous DDR approaches, both classical~\cite{nafchi2014phase} and learning-based~\cite{akbari2020binarization, lin2022three}, have successfully leveraged the spectral information of document images to enhance their visual quality. In recent works, the Discrete Wavelet Transform~\cite{mallat1989theory} has been used for Document Binarization to represent the document image in different frequency sub-bands, either to perform Segmentation~\cite{akbari2020binarization} or to obtain a ground truth ink map to train a Generative Adversarial Network~\cite{lin2022three}.
Our method exploits the Fourier Transform, localized only in frequency, to operate on the periodic structures common in the writing substrate. Unlike previous works, we do not use spectral information to represent the images but as an operator inside our proposed end-to-end Convolutional architecture.
In particular, we extend the FFC operator proposed by Chi~\etal~\cite{chi2020fast} to make it process volumetric data. 
Indeed, the FFC operator combines 2D convolutions and 2D discrete Fourier Transform (DFT)~\cite{cooley1965algorithm} and has been applied to spatial data for computer vision tasks such as inpainting~\cite{he2016deep}, super-resolution~\cite{sinha2022nl,zhang2022swinfir}, and semantic segmentation~\cite{berenguel2022fredsnet}. Some attempts have been made to apply the FFC to spatio-temporal data~\cite{chi2020fast}. Nonetheless, volumetric data are different from spatio-temporal ones, and thus, require ad hoc solutions~\cite{tran2015learning, qi2016volumetric, gonda2018parallel, panariello2022consistency}. Therefore, we argue that a solution based on spatio-temporal FFCs is not optimal for modeling the correlations between dimensions in volumetric data. In sight of this, we propose the volumetric FFC operator, which features 3D convolutions and 3D discrete Fourier Transform and better handles this kind of data.

\section{Proposed Approach}
\label{sec:method}
Our goal is to detect the presence of ink in the volumetric representation of a carbonized papyrus sheet. Our model takes as input such volumes and is expected to output an ink map with the same width and height of the volume, whose elements contain the probability of ink being in the corresponding papyrus surface. 
To tackle this task, we devise a U-net-like architecture, whose details are given in~\Cref{ssec:architecture}, featuring a variant of the FFC operator that we devise to handle volumetric data as described below.    

\subsection{Volumetric Fast Fourier Convolution}
The FFC proposed by Chi~\etal~\cite{chi2020fast}, which here we refer to as spatial FFC, is a neural operator that combines convolution and Fourier Transform to perform local reasoning in the space domain and non-local reasoning in the frequency domain. This information processing expands its receptive field and makes it suitable for handling pseudo-periodic patterns in the data. The operator has also been applied to spatio-temporal data~\cite{chi2020fast}. Here, we refer to this variant as Spatio-Temporal FFC (stFFC). The Volumetric FFC variant devised in this work presents the same properties of the spatial FFC operator but can directly handle volumetric data. A visual comparison between these mentioned operators is reported in Figure~\ref{fig:model}.

The vFFC operator takes as input a tensor $\mathbf{X} \in \mathbb{R}^{D \times H \times W \times 2C}$, where $2C$ is the number of channels, $D$ is the depth dimension, and $H$ and $W$ are the spatial dimensions.
This tensor is split into two parts along the channel dimension, which are fed to two interconnected branches: a local branch and a global branch. Splitting the input tensor into two chunks allows the encoding of different information in separate regions of the tensor. In addition, this approach permits the global and local branches to specialize in different aspects because they do not share the same inputs.
The \textit{local branch} contains two 3D convolutional layers with kernel size 3 and is in charge of modeling local volumetric information. 
In the \textit{global branch}, the input is mapped into the spectral domain to model global information. This branch consists of a 3D convolution with kernel size 3 and a 3D Spectral Transform (ST 3D) block. 

The ST 3D block exploits a three-dimensional real Fast Fourier Transform (FFT3D)~\cite{cooley1965algorithm}. Specifically, the FFT3D is performed across the depth, height, and width dimensions of the input tensor:
\begin{equation*}
        \text{FFT3D}(\mathbf{X}) = \mathbf{Z} = \mathbf{R} + i\mathbf{I},
\end{equation*}
where $\mathbf{Z}$ is a complex tensor, with real and imaginary parts $\mathbf{R}, \mathbf{I} \in \mathbb{R}^{D \times H \times \frac{W}{2} \times C}$. Then, $\mathbf{R}$ and $\mathbf{I}$ are stacked together along the channel axis, thus obtaining a tensor
\begin{equation*}
        \mathbf{R} || \mathbf{I} \in \mathbb{R}^{D \times H \times \frac{W}{2} \times 2C}
\end{equation*}
that is then fed to a $1{\times}1{\times}1$ convolutional layer. Batch normalization and ReLU activation functions are applied to the output of the convolution. The resulting tensor is reshaped into a complex-valued tensor $\mathbf{Z'} \in \mathbb{C}^{D \times H \times \frac{W}{2} \times C}$, and the inverse real Fast Fourier Transform (iFFT3D) is computed to obtain the global branch output:
\begin{equation*}
        \text{iFFT3D}(\mathbf{Z'}) = \mathbf{X'} \in \mathbb{R}^{D \times H \times W \times C}.
\end{equation*}

The output of each branch is summed to the output of the other, and the resulting tensors are fed to separate batch normalization and ReLU operation before being stacked together to obtain the final vFFC output tensor.

\subsection{Ink Detection Architecture}\label{ssec:architecture}
Rather than having a different response to X-rays than the papyrus support, the carbon-based ink in the Herculaneum scrolls ink modifies the substrate structural patterns in subtle ways, as shown in~\cite{parker2019invisibility}. In particular, parts containing ink have different types of cracks and densities and are thicker and smoother than empty ones. 
In this challenging case, it is important to consider both local and global contexts. The subtle patterns are, in fact, very localized, but with a global context, the model is able to consider holistically information from the whole subvolume. Moreover, we argue that the vFFC operator, which is able to handle pseudo-periodic patterns, is suitable for modeling this kind of data. 
We treat ink detection as a pixel-wise classification task, and we use an encoder-bottleneck-decoder architecture composed of a 3D convolutional encoder, a vFFC-based bottleneck, and a 2D convolutional decoder. 
To accurately localize the signal and fuse high-level and low-level features, we employ skip connections between the encoder and the decoder, similar to U-Net~\cite{ronneberger2015unet}, and vFFC layers in the latent space of the network. 

Specifically, we employ a 3D ResNet-34~\cite{he2016deep} to encode the subvolumes. The output from the last block is fed to 3 vFFC Residual Blocks, each combining two vFFC layers with a residual connection. The $\mathbb{R}^{D \times H \times W \times C}$ feature maps from the bottleneck are collapsed to $\mathbb{R}^{H \times W \times C}$ by averaging the depth dimension. The same applies to the encoder activations in the skip connections. Then, a 2D decoder reconstructs the ink map. This decoder is obtained by stacking four blocks made of 2-fold bilinear interpolations and convolutions, operating on the concatenation of the output of the previous layer and the features from the corresponding encoder layer. Since the fragments have very high resolution but represent small objects, the decoder does not reconstruct the full-scale ink map but rather a 4-fold downscaled one. This approach also limits the computational weight of the model.

\subsection{Training and Inference}
We train our model on subvolumes (3D patches) of the fragments scans both to limit the computational complexity of our model and to obtain more samples from the same fragment. Note that this strategy is customary for other data-constrained tasks such as document binarization~\cite{tensmeyer2017document} or medical images segmentation~\cite{kamnitsas2017efficient, de2018clinically}. Moreover, we apply a number of data augmentation operations (described in~\Cref{sec:experiments}) in order to reduce the risk of overfitting and help the model generalize to unseen data. Our adopted loss function is a combination of the Dice loss and weighted binary cross entropy (WBCE). The Dice score coefficient has been proposed for unbalanced segmentation~\cite{milletari2016v}, while the weighted binary cross-entropy is customary for classification. Let $N$ be the number of voxels, $p_i \in P$ the predicted binary segmentation map, and $g_i \in G$ the ground-truth ink image. The loss can be written as:  
\begin{equation*}
    \mathcal{L} = \text{Dice} + \text{WBCE},
\end{equation*}
with
\begin{equation*}
    \text{Dice} = \frac{2 \sum^N_n p_n g_n + \epsilon}{\sum^N_n p_n^2 + \sum^N_n g_n^2 + \epsilon},
\end{equation*}
\begin{equation*}
    \text{WBCE} = - \frac{1}{N} \sum^N_{n=1} w [g_n \log{p_n} + (1 -g_n) \log(1 - p_n)]
\end{equation*}
where $w$ is the weight attributed to the ink class.
The predicted ink map contains values ranging from 0 to 1, representing the ink presence probability. From this, we obtained a binarized image by applying a threshold of 0.5.
\section{Experimental Analysis}
\label{sec:experiments}
In this section, we present the experiment setup concerning the ink detection task on the fragments of the recently proposed EduceLab-Scrolls dataset and detail our training strategy. Moreover, we present an extensive analysis of the proposed components, both to explore their contribution and give some intuitions on what is more suitable for this task. 

\begin{figure}
    \centering
    \includegraphics[width=0.475\textwidth]{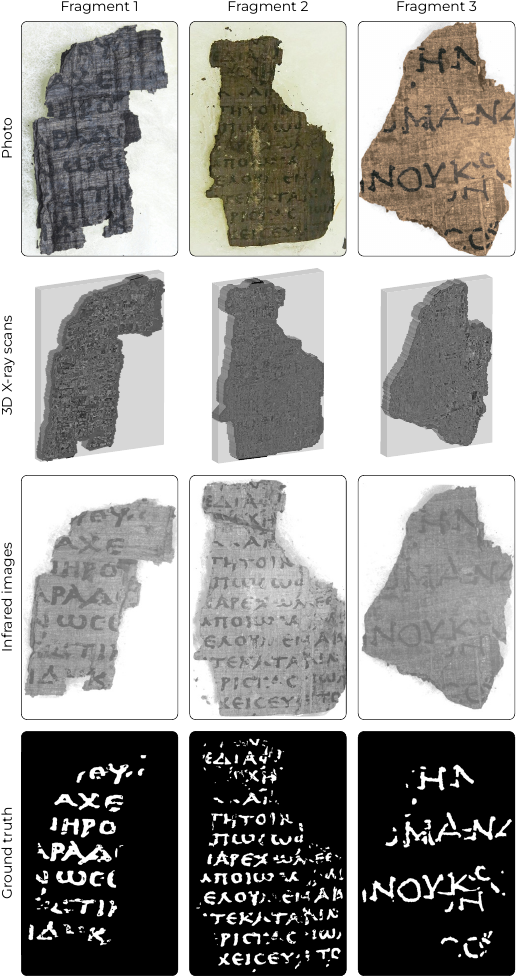}
    \caption{Fragments contained in the considered dataset, in different modalities.}
    \label{fig:dataset}
\end{figure}

\subsection{Experiment Setup}
\tit{Dataset} 
As mentioned above, in this work, we focus on the surface volumes of the fragments in the EduceLab-Scrolls~\cite{parsons2023educelab} (see Figure~\ref{fig:dataset}). These scans have been performed with X-ray micro-CT and thus are very detailed. With a 3.2$\mu m$ voxel size, they have resolutions in the order of thousands of pixels and weigh several Gygabites. In particular, we use the three publicly available fragments released for the Vesuvius Challenge on Ink Detection related to the EduceLab-Scrolls dataset project. These documents, broken and detached from the scrolls during destructive physical unrolling tentatives, have visible ink and, thus, it has been possible to obtain a ground truth ink map, also thanks to infrared analysis. 
The scans are all composed of 65 slices but have different spatial sizes: 6330$\times$8181 for fragment 1, 9506$\times$14830 for fragment 2, and 5249$\times$7606 pixels for fragment 3. In our experiments, we use fragment 1 for test and the other two for training. 

\tit{Evaluation Metrics}
Considering the novelty of the task and its similarities to other tasks, we use a combination of metrics to evaluate our models. First of all, we adopt the score proposed in~\cite{parsons2023educelab} to evaluate the ink detection performance, \ie~the F$_\beta$ score, which weighs more the precision than the recall and is defined as follows:
\begin{equation*}
   \text{F}_\beta = \frac{(1 + \beta^2)pr}{\beta^2p+r},  
\end{equation*}
where $p$ and $r$ are precision and recall, respectively, and the parameter $\beta{=}0.5$.
Moreover, we employ some scores commonly used in Document Binarization, namely the pseudo-FMeasure (pFM)~\cite{ntirogiannis2012performance} and the Peak Signal-to-Noise Ratio (PSNR), which measures the similarity of two images.

\tit{Implementation Details}
Considering that most of the information is in the central slices and to reduce computational impact, we use 16 slices for training. Note that using more slices would not straightforwardly improve performance: in fact, fragments have hidden text, in papyrus layers fused in the back, that could be present in the surface slice if we go too deep. Taking also into account that the ink is a localized signal, we train on 16$\times$256$\times$256 subvolumes. 

For optimization, we use the AdamW~\cite{loshchilov2017decoupled} optimizer with $\beta_1=0.9$, $\beta_2=0.95$, and the OneCycle scheduler~\cite{smith2019super} with learning rate 0.003. We set the batch size to 4 and apply floating point 16 mixed precision to speed up training. For regularization, we use Drop Path~\cite{larsson2016fractalnet} with rate 0.1, Channel Dropout~\cite{tompson2015efficient}, with rate and maximum dropped channels both set to 0.5, and additional data augmentation strategies as described in~\Cref{subsection:data_augmentation}. 
We set the ink weight in the WBCE to 1 and train the models for 20 epochs, which takes 24 hours with an NVIDIA RTX A5000. 
At test time, for the entire test fragment, we perform the prediction on 16$\times$256$\times$256 subvolumes and keep only the prediction on the inner 16$\times$128$\times$128 part to reduce border artifacts. 

\tit{Data Augmentation}
\label{subsection:data_augmentation}
We perform data augmentation in order to enhance model generalization and robustness. 
In particular, we obtain the training subvolumes starting from a 3D lattice of the whole fragment scans, with cells of size 32$\times$512$\times$512 and stride 64. During each training iteration, we extract subvolumes of size 16$\times$256$\times$256 from a random 3D position of the lattice cells. The rationale behind this strategy is twofold. In the spatial dimension, the random crop increases the variability in the training data, preventing the network from being exposed to the same patches repetitively. In the depth dimension, exposing the network to different slices ensures that the model learns to discern patterns across all depth coordinates. 
Indeed in a real-world application, the surface volumes would be extracted from segmented sheets in a scroll. This process is prone to errors and misalignment. Thus, there is rarely correspondence between depth coordinates in different samples. By enforcing depth invariance, the network is able to model patterns independently of specific depth locations. 
Moreover, we randomly apply horizontal and vertical flip, 90$^\circ$ random rotation, and transposition to obtain the corresponding Dihedral group $D_4$ transformations, further increasing diversity in spatial patterns. 

\begin{figure*}[t]
    \centering
    \includegraphics[width=\textwidth]{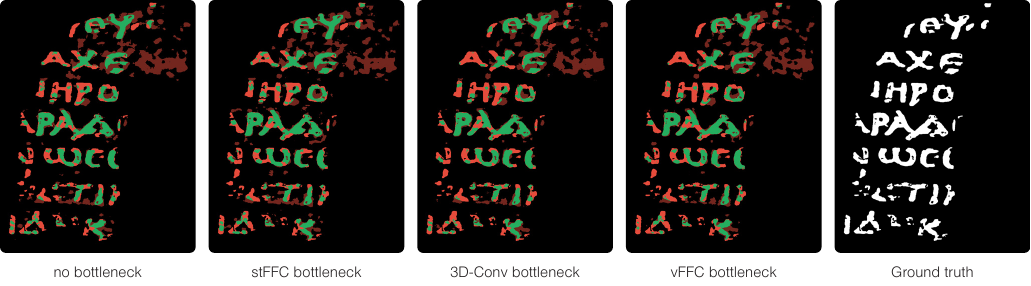}
    \caption{Qualitative results of the considered variants on the test fragment. Red indicates the missed ink predictions, green indicates the pixels correctly identified as ink, and maroon indicates the background pixels incorrectly classified as ink.}
    \label{fig:qualitatives}
\end{figure*}

\subsection{Results}
In this section, we report an analysis aimed at identifying the most relevant elements of the training procedure and assessing the effectiveness of our proposed vFFC-based model for the ink detection task.

\tit{Training Startegy}
First, we analyze the contribution of the applied data augmentation strategies. The results of this analysis are reported in~\Cref{tab:transformations}. It can be observed that the Dihedral transformations bring the most benefits to the training by inducing the larger variability. From these experiments on fragment 1, the benefit of the random crop is not evident. Nonetheless, when testing on the public and the private test set of the Kaggle Ink Detection Challenge associated with the EduceLab-Scrolls dataset, this regularization strategy has been proven beneficial for increasing the generalization capability of our model in handling fragments in which the relevant information is localized in different slices. The effect of the random crop, especially on the depth axis, is visible from the activation maps discussed in~\Cref{ssec:maps}.
Then, we perform an ablation on the training loss terms, whose results are reported in~\Cref{tab:losses}. It emerges that the WBCE leads to good performance, and the combination with the Dice helps refine the results. Moreover, giving more weight to the ink class leads to worse performance, despite the ink class being much less represented than the background. 

\begin{table}[t]
    \begin{center}
    \setlength{\tabcolsep}{.4em}
    \resizebox{\linewidth}{!}{
    \begin{tabular}{c c c c c c c}
    \toprule
            \makecell{\textbf{Dihedral} \\ \textbf{Transform}} & \makecell{\textbf{Random} \\ \textbf{Crop}} & \makecell{\textbf{Channel}\\ \textbf{Dropout}} & \textbf{F$_\beta$} & \textbf{pFM} & \textbf{PSNR} \\
    \midrule
    -           & -          & -          & 0.37 & 0.53 &  9.37 \\ 
    \checkmark  & -          & -          & 0.46 & 0.57 & \textbf{10.38} \\
    \checkmark  & \checkmark & -          & 0.46 & 0.57 &  9.62 \\ 
    \checkmark  & \checkmark & \checkmark & \textbf{0.47} & \textbf{0.58} & 10.09 \\ 
    \bottomrule
    \end{tabular}
    }
    \end{center}
    \caption{Ablation analysis on the augmentation strategies adopted.}
    \label{tab:transformations}
\end{table}

\begin{table}[t]
    \begin{center}
    \resizebox{0.85\linewidth}{!}{
    \begin{tabular}{l c c c c c c}
    \toprule
\textbf{Loss} & \textbf{$w$} & \textbf{F$_\beta$} & \textbf{pFM} & \textbf{PSNR}\\
    \midrule
    Dice  & - & 0.40 & 0.56 &  9.47    \\ 
    WBCE & 1 & 0.44 & 0.56 & \textbf{10.34} \\ 
    WBCE + Dice  & 5 & 0.41 & 0.53 &  8.20  \\ 
    WBCE + Dice  & 2 & 0.45 & 0.56 &  9.86  \\ 
    WBCE + Dice  & 1 & \textbf{0.47} & \textbf{0.58} & 10.09   \\ 
    \bottomrule
    \end{tabular}
    }
    \end{center}
    \caption{Ablation analysis on the training loss function.}
    \label{tab:losses}
\end{table}

\tit{Architecture} 
We compare the proposed architecture with a number of baselines in \Cref{tab:arch_ablation}. To assess the effectiveness of vFFCs, we compare our proposed model against variants featuring different kinds of bottlenecks. In particular, we consider a variant without bottleneck layers, one with stFFCs, and one with 3D-Convs. Overall, using the proposed vFFC in the bottleneck leads to the best performance. Arguably, this is due to the more precise prediction of the background pixels, as can also be observed from the qualitative comparison in~\Cref{fig:qualitatives} (see, \eg~the top-right part of the prediction map).

\begin{table}[t]
    \begin{center}
    \resizebox{0.7\linewidth}{!}{
    \begin{tabular}{l c c c c }
    \toprule
    \textbf{Bottleneck} & \textbf{F$_\beta$} & \textbf{pFM} & \textbf{PSNR} \\
    \midrule
    -      & 0.46 & \textbf{0.58} &  9.87 \\  
    stFFC  & 0.45 & \textbf{0.58} &  9.76 \\  
    3D-Conv & 0.46 & \textbf{0.58} & 10.04 \\ 
    vFFC   & \textbf{0.47} & \textbf{0.58} & \textbf{10.09} \\ 
    \bottomrule
    \end{tabular}
    }
    \end{center}
    \caption{Quantitative comparison between our model featuring vFFCs in the bottleneck and baselines with different bottlenecks.}
    \label{tab:arch_ablation}
\end{table}

\begin{figure}[t]
    \centering
    \includegraphics[width=\linewidth]{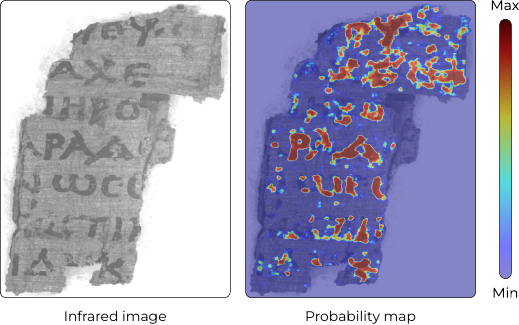}
    \caption{Prediction probability map of our proposed model on the test fragment.}
    \label{fig:prediction_map}
\end{figure}

\tit{Kaggle Ink Detection Challenge Results}
We participate in the Kaggle Vesuvius Challenge - Ink Detection\footnote{www.kaggle.com/competitions/vesuvius-challenge-ink-detection} with a boosted version of our model. In particular, we maintain the architecture but employ a different training and inference strategy to fully exploit the training fragments available. In particular, we split each available fragment into two parts in order to obtain a total of six subsets. Then, we train six versions of our model in a k-fold strategy and set the prediction threshold to 0.8. We then submit an ensemble of these models, combined via majority voting on each pixel with four votes out of six. The submission scored silver medal (\ie~within the top-5\% submissions) in the competition, scoring around F$_\beta$=0.70 on the public test set and F$_\beta$=0.60 on the private one.

\subsubsection{Visualizations}\label{ssec:maps}
For reference, in~\Cref{fig:prediction_map}, we report the ink prediction probability of our model. Although we set the prediction threshold to 0.5, we argue that such a non-thresholded visualization can be helpful when visually inspecting the fragments.

\begin{figure}
    \centering
    \includegraphics[width=\linewidth]{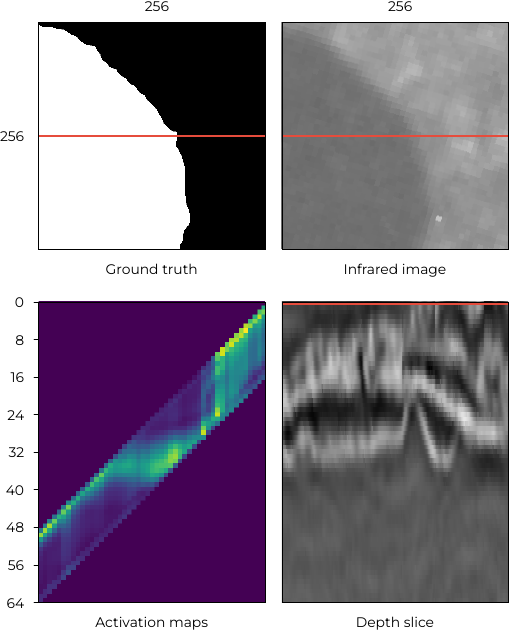}
    \caption{Activation maps by varying the depth and keeping the spatial coordinates of the subvolume fed to our model (blue represents the minimum value, yellow the maximum). For reference, we report the corresponding ground-truth ink map, the infrared image of the surface, and the coronal slice of the subvolume.}
    \label{fig:activations}
\end{figure}

Finally, we study the model class activation map by using LayerCam~\cite{jiang2021layercam}. In particular, we consider the whole depth size for a given 2D patch on the papyrus surface, obtaining a volume $\mathbf{V}$ with shape $65 \times 256 \times 256$. Then, we extract subvolumes of size $\mathbf{v} \subset \mathbf{V}$ with shape $16 \times 256 \times 256$ and stride set to 1, covering the whole $Z$ axis, and feed them to the network. We compute the mean activation value on the spatial dimensions for each depth slice and show the result in~\Cref{fig:activations}. On the x-axis of the activation maps plot, we report the value of the starting $z \in Z$ coordinate of the subvolume. On the y-axis, we plot the absolute $z \in Z$ depth coordinate. As we can see, the plot is divided into three regions, corresponding to the upper ($0 \leq z < 24$), middle ($24 \leq z < 40$), and lower ($40 \leq z \leq 65$) parts of the papyrus volume $\mathbf{V}$. 
When we feed to the model slices corresponding to small values of $z$, they are all informative; thus, the network does not look specifically at any of them but rather combines the patterns. In the center-most region of the plot, the network consistently focuses on the slice with $z \approx 30$ for several slidings of the subvolume. We argue that this is the main manifestation of the depth invariance that we instill with the subvolume random crop. This way, the network can account for segmentation misalignment in real-world scenarios, learning to recognize important slices independently from their relative position in the subvolume. Then, with growing values of $z$, the depth slices contain less and less information, so the network focuses on the slices with relatively lower values of $z$. We provide visualizations of the volume $\mathbf{V}$ in~\Cref{fig:activations}: from the depth slice, we can see that the information is for the most part in the low $z$ coordinates, while the infrared image and the ground truth show where ink is present in the surface and changes the inner texture of the papyrus support.

\section{Conclusion and Future Work}
\label{sec:conclusion}
In this work, we have devised the vFFC operator, a modification of the standard FFC designed to handle volumetric data directly. Moreover, we have proposed to incorporate the vFFC in an architecture for the DDR sub-task of ink detection on volumetric scans of carbonized papyri fragments from the EduceLab-Scrolls dataset. Through experimental analysis, we have assessed the effectiveness of our approach and hopefully contribute to the emerging research interest in DDR on the challenging Herculaneum papyri data.
Finally, we argue that the vFFC operator could also be applied to other tasks and scenarios involving volumetric data (\eg~medical imaging), which we leave for future work.

\section*{Acknowledgement}
This work was supported by the ``AI for Digital Humanities'' project (Pratica Sime n.2018.0390), funded by ``Fondazione di Modena'' and the PNRR project Italian Strengthening of Esfri RI Resilience (ITSERR) funded by the European Union – NextGenerationEU (CUP: B53C22001770006).

{\small
\balance
\bibliographystyle{ieee_fullname}
\bibliography{main}
}

\end{document}